\documentclass[preprint,conference]{IEEEtran}
\IEEEoverridecommandlockouts
\usepackage{cite}
\usepackage{amsmath,amssymb,amsfonts}
\usepackage{algorithmic}
\usepackage{graphicx}
\usepackage{textcomp}
\usepackage{xcolor}

\def\BibTeX{{\rm B\kern-.05em{\sc i\kern-.025em b}\kern-.08em
    T\kern-.1667em\lower.7ex\hbox{E}\kern-.125emX}}

\usepackage{stfloats}
\usepackage{booktabs}
\usepackage{multirow}
\usepackage{longtable}
\usepackage{amsmath}
\usepackage[inline]{enumitem}

\usepackage{balance}

\IEEEoverridecommandlockouts

\usepackage{tikz}
\usepackage{textcomp}
\usepackage{hyperref}
\usepackage{lipsum}

\newcommand\copyrighttext{%
  \footnotesize \textcopyright 2025 IEEE. Personal use of this material is permitted. Permission from IEEE must be obtained for all other uses, in any current or future  media, including reprinting/republishing this material for advertising or promotional  purposes, creating new collective works, for resale or redistribution to servers or lists, or reuse of any copyrighted component of this work in other works. \\ This is a preprint of the accepted paper at IEEE CogMI 2025 - The 7th IEEE International Conference on Cognitive Machine Intelligence.

}
\newcommand\copyrightnotice{%
\begin{tikzpicture}[remember picture,overlay]
\node[anchor=south,yshift=10pt] at (current page.south) {\fbox{\parbox{\dimexpr\textwidth-\fboxsep-\fboxrule\relax}{\copyrighttext}}};
\end{tikzpicture}%
}

\begin{document}

\title{
Knowledge-guided Continual Learning\\  
for Behavioral  Analytics Systems 
}

\author{\IEEEauthorblockN{Yasas Senarath}
\IEEEauthorblockA{
\textit{Humanitarian Informatics Lab} \\ 
\textit{George Mason University}\\
Fairfax, VA, USA \\
ywijesu@gmu.edu}
\and
\IEEEauthorblockN{Hemant Purohit}
\IEEEauthorblockA{
\textit{Humanitarian Informatics Lab} \\
\textit{George Mason University} \\
Fairfax, VA, USA \\
hpurohit@gmu.edu}
}

\maketitle

\copyrightnotice

\begin{abstract}
User behavior on online platforms is evolving, reflecting real-world changes in how people post, whether it's helpful messages or hate speech. Models that learn to capture this content can experience a decrease in performance over time due to data drift, which can lead to ineffective behavioral analytics systems. However, fine-tuning such a model over time with new data can be detrimental due to catastrophic forgetting. Replay-based approaches in continual learning offer a simple yet efficient method to update such models, minimizing forgetting by maintaining a buffer of important training instances from past learned tasks. However, the main limitation of this approach is the fixed size of the buffer. External knowledge bases can be utilized to overcome this limitation through data augmentation. We propose a novel augmentation-based approach to incorporate external knowledge in the replay-based continual learning framework. We evaluate several strategies with three datasets from prior studies related to deviant behavior classification to assess the integration of external knowledge in continual learning and demonstrate that augmentation helps outperform baseline replay-based approaches.
\end{abstract}

\begin{IEEEkeywords}
lifelong machine learning, deviant behavior classification, knowledge-guided learning, Wiktionary
\end{IEEEkeywords}

\section{Introduction}
\label{sec:introduction}

\textcolor{red}{Note: This paper includes offensive and hateful language. We have deliberately included these terms to ensure clarity and support efforts to counter hate speech.}

Online platforms support various forms of user engagement, including sharing text and images, commenting on posts, and resharing or liking content. This user-generated content often holds significant value for behavioral analytics in diverse application domains. For example, during disasters, individuals have leveraged these platforms to request help and coordinate relief efforts~\cite{purohit2020social}. However, deviant actors leverage these platforms to propagate inappropriate content that could induce violence, affecting social cohesion in a community~\cite{carley2020social,purohit2018intent}. Detecting such content (e.g., hate speech~\cite{yoder2022hate} and fear speech~\cite{saha2023rise}) provides the means to mitigate its reach. Platform moderators typically perform this task with the help of automated tools that utilize state-of-the-art deep learning models for text-based behavior  classification. 

Changes in real-world events commonly influence the nature of online discussions and such malicious behavior. For example, the COVID-19 pandemic led to the prevalence of different forms of negative discourse over time (e.g., rise in anti-Asian hate speech and anti-vaccine sentiment)~\cite{criss2021advocacy}. This dynamic nature of online content, coupled with the intent of deviant actors to increase their reach by evading detection mechanisms, can induce conceptual drift~\cite{lu2018learning}. Consequently, statistically learned behavior classification models (like fine-tuned BERT models) may experience performance degradation over time~\cite{qian2021lifelong}.
Systems using such models can mitigate this effect by continually training or fine-tuning a model with new data. However, updating existing models in this manner presents several challenges. First, continual learning (CL) causes catastrophic forgetting~\cite{mccloskey1989catastrophic}, where an existing model trained on a dataset from a new domain or task may experience a decrease in performance on previously learned tasks. Second, models that are continually learned often exhibit lower performance compared to models trained on the entire combined dataset~\cite{qian2021lifelong}. However, maintaining the full dataset and training are both computationally expensive and sometimes impractical due to concerns about data privacy. 

\begin{figure}
    \centering
    \includegraphics[width=1\linewidth]{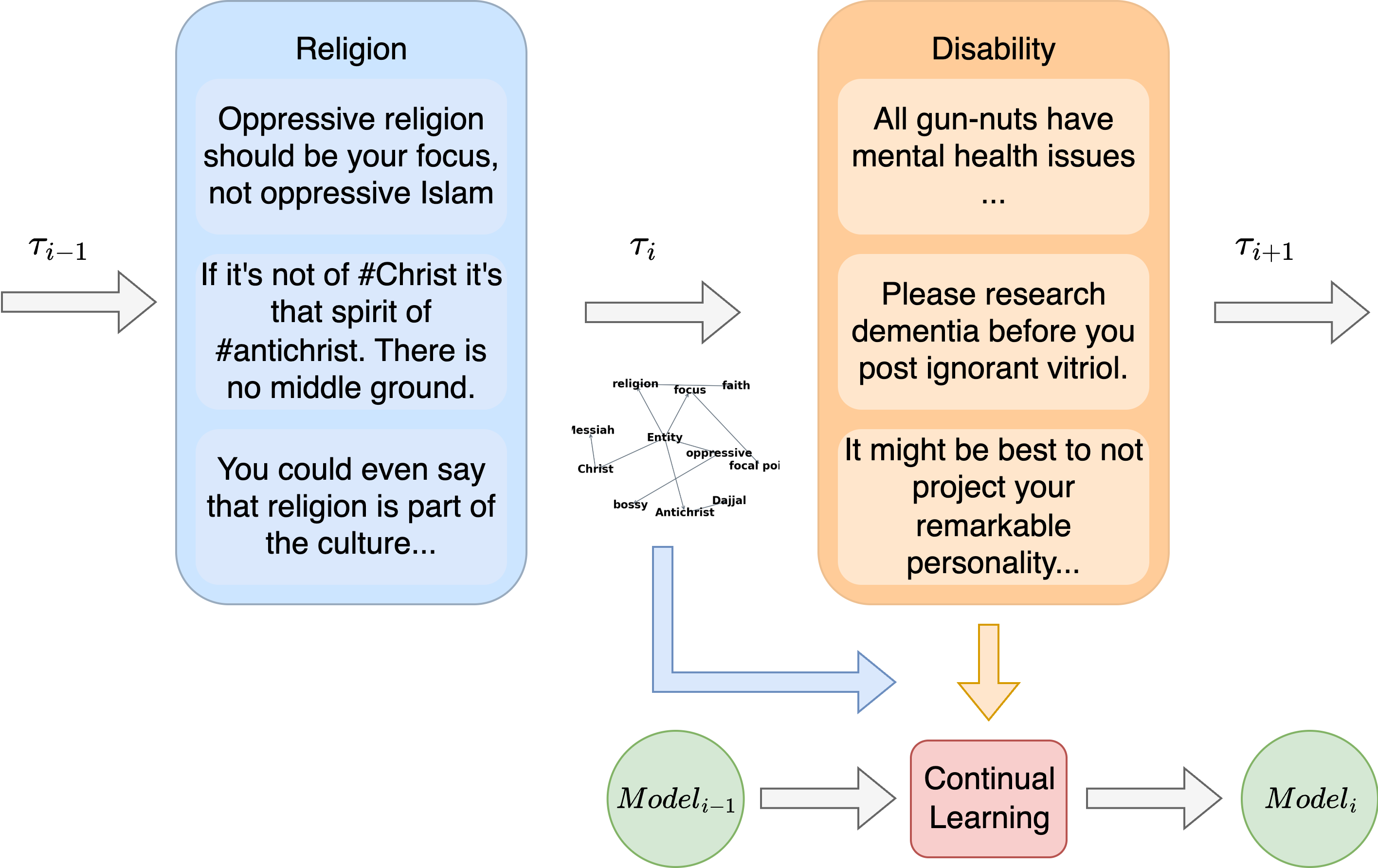}
    \caption{An illustration of a task series from an existing dataset~\cite{kennedy2020constructing} and an overview of the proposed model training with an external knowledge graph. 
    }
    \label{fig:intro-example}
\end{figure}

Prior research has examined different methods 
to address the  
aforementioned  challenges for CL~\cite{kirkpatrick2017overcoming}. A prevalent CL method in both natural language processing and computer vision is the replay-based approach, which involves using a memory buffer to preserve a small subset of training data from each learning task. This retained data is subsequently used during the training of the model on new data. Various methods have been explored to optimize the utilization of this previously acquired knowledge, such as Gradient of Episodic Memory (GEM)~\cite{lopez2017gradient}.

This paper focuses on improving existing CL methods, particularly replay-based CL, which is developed to mitigate catastrophic forgetting through exemplar replay in the domain of deviant behavior classification. Exemplars are a subset of the original task's training data, typically selected to be representative or critical of the task. Previous research employs various strategies for exemplar selection. Prominent methods include random selection and cluster-based approaches~\cite{qian2021lifelong,lopez2017gradient}. Additionally, strategies that leverage the properties of the learned model for a given task have been explored. The premise of our study is that external knowledge helps to mitigate the limitations of the exemplar set in two distinct ways. \textit{First}, we enhance the exemplar selection process by augmenting the training data from the current task and subsequently selecting exemplars from the combined original and augmented datasets using a clustering-based approach. \textit{Second}, these selected exemplars are then incorporated into the learning process after undergoing a re-augmentation step. This procedure effectively increases the size of the training dataset for previous tasks, thereby improving the overall learning process. Figure~\ref{fig:intro-example} shows this process with an illustrative task series. 

This paper addresses the following research questions: 

\begin{enumerate}[start=1,label={\bfseries RQ\arabic*:},leftmargin=*,labelwidth=2cm, align=left]
\item How can we integrate external knowledge through augmentation in replay-based continual learning to reduce catastrophic forgetting (i.e., negative backward transfer) and enhance overall system performance? 
\item  In replay-based continual learning, should external knowledge be integrated during memory sampling stages, data replay stages, or both? 
\end{enumerate}

In addressing the above questions, we make the following contributions: \textbf{(1)} We introduce an extensible learning framework that facilitates the integration of external knowledge into the CL process for text classification tasks. Our approach differs from previous studies that primarily focus on leveraging internal knowledge, such as using the extracted knowledge from previous tasks~\cite{zhou2024class}. \textbf{(2)} Our proposed knowledge-guided CL framework addresses a major challenge of evolving human behavior for online behavioral analytics systems and has applications across a wide range of domains. \textbf{(3)} We conduct an extensive evaluation of the proposed method, including an ablation study to quantify the impact of each component of the framework and to identify the optimal settings. \textbf{(4)} 
To enable future research and ensure reproducibility, we publicly release all resources used in the experiments, including augmented knowledge bases, models used in the CL process, source code, and processed datasets
~\footnote{https://github.com/ysenarath/CogMI-2025-KG-CL}.

The remainder of this paper is organized as follows. Section~\ref{sec:related-works} reviews the related work, Section~\ref{sec:problem-formulation} formally introduces the problem along with a background, and the methodology is presented in Section~\ref{sec:method}. Section~\ref{sec:experiments} describes the experimental setup, including datasets used for the empirical validation of the proposed approach. Section~\ref{sec:results} discusses the results of our research questions accordingly. Finally, Section~\ref{sec:conclusion} concludes the paper with future directions.

\section{Related Work}
\label{sec:related-works}

\subsection{Continual Learning (CL) Approaches}

Methods for CL, also known as lifelong learning~\cite{chen2018lifelong},  incremental learning~\cite{solomonoff1989system},  sequential learning~\cite{mccloskey1989catastrophic}, and never-ending learning~\cite{carlson2010toward}, have been explored in various domains. The goal of CL methods is to learn a sequence of tasks incrementally, utilizing previously acquired knowledge to enhance learning on new tasks while mitigating the risk of catastrophic forgetting~\cite{chen2018lifelong}. Our research aims to mitigate the risk of catastrophic forgetting by leveraging external knowledge. 

Three primary CL paradigms have been studied within the academic literature: task-incremental learning (Task-IL), domain-incremental learning (Domain-IL), and class-incremental learning (Class-IL). In Domain-IL, the class labels remain consistent across all tasks. Task-IL is characterized by the use of explicit task identifiers during the inference phase. In contrast, Class-IL introduces new classes sequentially in each task, and the task identifier is not disclosed during inference. This study focuses on the Class-IL problem. 

Previous research has proposed different approaches to address CL challenges. One prominent class of methods, known as parameter-regularization-based approaches, operates by incorporating an additional penalty term into the loss function. This penalty discourages changes to parameters that are critical for previously learned tasks, thereby mitigating the problem of catastrophic forgetting. A well-known example of this approach is Elastic Weight Consolidation (EWC) ~\cite{kirkpatrick2017overcoming}. Distillation is another regularization-based approach wherein a model trained on a new task incorporates distilled knowledge from the model of a previous task~\cite {monaikul2021continual,li2022continual}. A major limitation of the distillation approach is its vulnerability to domain shifts~\cite{tang2024direct}. Methods based on parameter isolation and dynamic architecture are also explored in CL. They use a unique set of parameters in learning a task, either by preventing parameter updates through masking during learning or by dynamically adding or updating neural network architecture~\cite{rusu2016progressive,wortsman2020supermasks}. Replay-based methods are another set of important CL approaches. These methods usually either keep a small set of training examples (called exemplars) from previous tasks in a buffer to use during new learning~\cite{rebuffi2017icarl,qian2021lifelong} or use a generator to create pseudo-exemplars of previous tasks~\cite{kemker2018fearnet}. Typically, learning is performed using replay, where the buffer data are combined with the training data for the current task and optimized jointly. Data regularization techniques, such as Gradient Episodic Memory (GEM)~\cite{lopez2017gradient} and Averaged Gradient Episodic Memory (A-GEM)~\cite{AGEM}, have been proposed to address overfitting associated with the replay-based approach.

The objective of this study is to improve replay-based methods by incorporating external knowledge. This approach addresses the primary limitations of existing methods, such as the restricted buffer size, which is often a consequence of resource constraints inherent to CL scenarios and privacy concerns related to data retention. Additionally, the literature provides a limited exploration of such CL methods in domains of online behavioral analytics~\cite{qian2021lifelong,omrani2024towards}.

\subsection{Knowledge-based Augmentation in Text Classification}

%
Knowledge bases such as Wikipedia and WordNet have been leveraged in multiple ways to enhance text classification, including integrating knowledge as features~\cite{senarath2020evaluating} and employing it for data augmentation~\cite{bayer2022survey}.
Among these approaches, knowledge-based augmentation has been widely adopted for integrating external knowledge, for its simplicity.
A systematic review conducted by Bayer et al. \cite{bayer2022survey} outlines various methodologies employed for augmentation. The review demonstrates that knowledge-based augmentation can be implemented at multiple levels, including character, word, phrase, or document level, in either the feature space or data space. Among these, synonym replacement has emerged as a particularly prevalent augmentation strategy attributed to its simplicity and effectiveness in enhancing classification performance~\cite{bayer2022survey}. 

In this study, we adopt a similar strategy to synonym replacement as our primary augmentation technique because of these advantages. However, a significant limitation of this approach, particularly in the context of detecting online deviant behavior, lies in the inadequacy of conventional general knowledge bases, such as WordNet, to encapsulate the semantics of deviant behaviors in online user-generated content. Consequently, existing research that predominantly relies on such resources may not effectively capture the nuanced linguistic characteristics associated with behaviors such as posting hate speech, and it is not updated frequently. Therefore, we posit that the use of a generic crowdsourced knowledge base can help CL. We explore how such a knowledge base can be effectively utilized within a CL framework.

\section{Problem Formulation}
\label{sec:problem-formulation}

The overarching problem addressed in this paper is framed as a text classification task in a Class-IL setting. While text classification has been extensively studied in the literature, the challenge of learning multiple tasks in a continual manner has received comparatively less attention. 
In this section, we formulate our novel problem as a continual text classification task that integrates external knowledge into the continual learning process. This formulation is broadly applicable to a wide range of text classification problems. 

\noindent \textbf{\underline{Preliminaries}: } Let us assume that there is a user-generated text instance $x_{t,i}$ corresponding to the task $\tau_t$, where $i$ denotes a specific instance. The objective is
multi-class classification, i.e., to classify a given instance across a set of classes observed up to and including the current task $\tau_L$, without having access to the task information of the instance. This problem can be formalized using the following function:

\begin{equation}
\hat{y}_{t,i} = \arg\max_{c \in \mathcal{C}_{\leq L}} \; p(c \mid x_{t,i})
\label{eqn:inference-simple}
\end{equation}

where $\mathcal{C}_{\leq L} = \bigcup_{k=0}^{L} \mathcal{C}_k$ denotes the union of all class labels observed in tasks $\tau_0$ through $\tau_L$ and $\mathcal{C}_k$ indicates classes of the task $\tau_k$.

In our study, for simplicity, we define a task as a domain to which the data instance belongs. For example, in hate speech classification, an instance may belong to a domain based on the target of the hate speech (e.g., by age, by gender, etc.). When formulating the dataset for this problem (training and testing data), we derive the labels such that they are not shared across tasks. It means if the same label (e.g., `hateful') is observed in multiple tasks, they will be treated as two different classes (e.g., age:hateful and gender:hateful). This formulation can be classified as the Class-IL setting in the continual learning literature.

\noindent \textbf{\underline{Problem Statement}: } 
Given a new task $\tau_{t}$, its associated training data $\mathcal{D}^{\text{train}}_{t}$, a neural network-based model trained up to the task $\tau_{t-1}$, the accumulated internal knowledge up to that task $\mathcal{K}^{\text{int}}_{t-1}$, and an external knowledge base $\mathcal{K}^{\text{ext}}$, the objective is to minimize classification loss in the training data for the new task. This optimization is performed using task-specific data and instances derived from both internal and external knowledge sources, as specified in Equation~\ref{eqn:continual-replay-full}. This problem formulation closely resembles that of standard direct replay-based methods in the CL literature, but it incorporates generalizations to accommodate external knowledge. 

\begin{align}
\theta_t = \arg\min_{\theta} \; \Bigg[
& \frac{1}{|\mathcal{D}_t^{\text{train}}|} \sum_{(x, y) \in \mathcal{D}_t^{\text{train}}} \mathcal{L}(f_\theta(x), y) \nonumber \\
& + \lambda \cdot \frac{1}{|\mathcal{D}_t^{\text{replay}}|} \sum_{(x, y) \in \mathcal{D}_t^{\text{replay}}} \mathcal{L}(f_\theta(x), y)
\Bigg]
\label{eqn:continual-replay-full}
\end{align}

where $\mathcal{D}_t^{\text{replay}}$ is the additional data provided from internal and external knowledge obtained from Equation~\ref{eqn:replay-data-generator}, and $\lambda$ controls the relative weight of the replay loss. $f_\theta(x)$ is the neural network with parameters $\theta$ (weights and biases), and x is the input data.

\begin{equation}
\mathcal{D}_{t}^{\text{replay}} = g(\mathcal{K}^{\text{int}}_{t-1}, \mathcal{K}^{\text{ext}})
\label{eqn:replay-data-generator}
\end{equation}

The replay data is generated using Equation~\ref{eqn:replay-data-generator}, where $g$ is a function that incorporates both internal and external knowledge to construct the replay dataset $\mathcal{D}_t^{\text{replay}}$. Specific details of the function $g$ are described in the methodology, Section~\ref{sec:augmentation}.

The function $g$ is responsible for generating data that supports learning a new task without forgetting the previous ones. However, for this function to work effectively, it requires access to data from past tasks, which must be stored in a buffer. Therefore, it is essential to strategically store instances in the buffer to make efficient use of them during the data replay process. In this research, internal knowledge consists solely of a fixed-size buffer, consistent with existing work in the CL literature~\cite{qian2021lifelong}. Let the maximum size of the buffer be $M$ instances. When a new task $\tau_{t}$ is introduced, a selection process is performed both within the buffer and from the training data for the new task. This ensures that each task is assigned an equal number of instances, specifically $M / t$. Various techniques are used for data selection, including random sampling, stratified random sampling, and cluster-based methods for defining the selection function. More details on these data selection settings are provided in Section~\ref{sec:experiment-settings}. 

\section{Methodology}
\label{sec:method}

\begin{figure*}[!ht]
    \centering
    \includegraphics[width=0.6\linewidth]{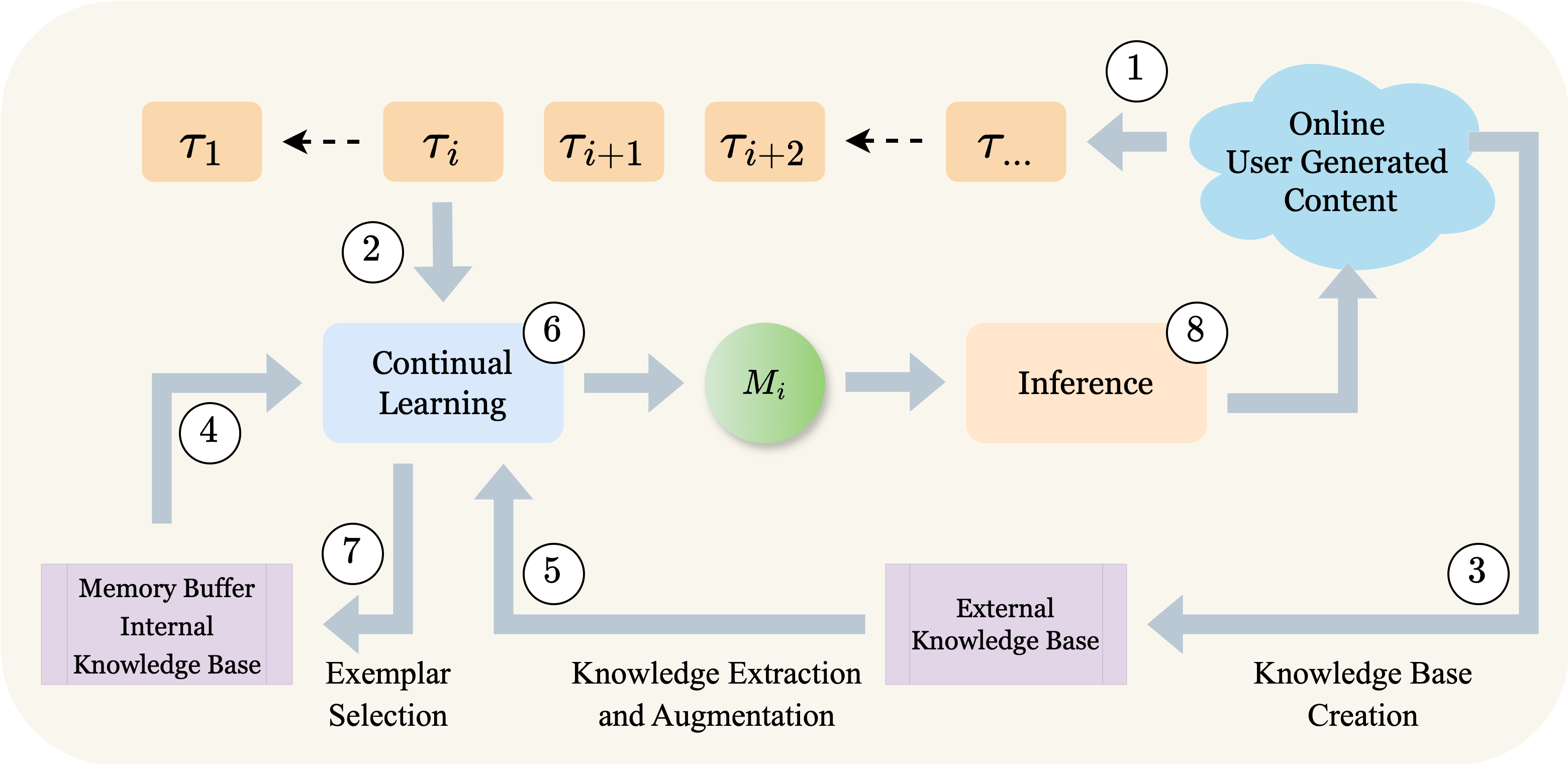}
    \caption{The proposed knowledge-guided CL system that incorporates external knowledge in the learning process. The numbers in circles indicate the order in which the data flows, as described in Section~\ref{sec:method}.} 
    \label{fig:cl-system}
\end{figure*}

In this section, we describe key components of the proposed knowledge-guided CL framework (see Fig.~\ref{fig:cl-system}): CL system, Knowledge Base Creation, Knowledge Extraction \& Augmentation, and Memory Buffer \& Exemplar Selection.   

\subsection{Continual Learning (CL) System}
\label{sec:method-cl-system}

In the proposed system, we enhance the traditional replay-based CL approach by utilizing external knowledge to improve the overall system performance. Consistent with previous research, we use a fixed-size memory buffer (internal knowledge) to store instances and fine-tune the model on the training datasets of the new tasks as they become available. Figure~\ref{fig:cl-system} illustrates the high-level architecture of the modified framework for the CL process with the proposed integration of knowledge bases. First, the data from the online user-generated content is divided into tasks~\textcircled{\textbf{1}}, and then each new task is presented for training~\textcircled{\textbf{2}}. We assume each task has training data with human-annotated labels related to that task. The next step after observing a new task is to update the external knowledge base to the latest version by downloading required resources and preprocessing as detailed in Section~\ref{sec:knowledge-base-creation}~\textcircled{\textbf{3}}. 

During the first task, the CL system learner \textcircled{6} initializes a pre-trained transformer model as the backbone for text encoding with an additional dense layer for classification. The classification layer uses a softmax activation to produce probability distributions over all classes in the first task. In this step, neither external nor internal knowledge bases are used during fine-tuning; the model is fine-tuned solely to classify the labels introduced in the first task. After updating the weights of the model to accommodate the first task, the CL system learner updates the memory buffer with the help of the external knowledge module. We discuss this process in detail in Section~\ref{sec:method-exempar-selection}.

In subsequent tasks, the CL system learner extends the output layer of the model from the previous task to support the additional classes by adding new classification nodes while keeping the weights and biases for the nodes related to the classes from previous tasks unchanged. Then, the CL system learner takes data related to the new task and the data in the buffer~\textcircled{4} to fine-tune the existing model ($M_{i-1}$) with the help of the external knowledge base~\textcircled{5}. The external knowledge base is used to augment the exemplars in the buffer using several strategies before being used to supplement the current training data. More details on the augmentation process and the strategies will be provided in Section~\ref{sec:augmentation}. Like in the first step, the memory buffer will be updated after learning about the current task~\textcircled{7}. Finally, the learned model will be used for inference until a new task is observed~\textcircled{8}.

\begin{figure*}[!ht]
    \centering
    \includegraphics[width=0.98\linewidth]{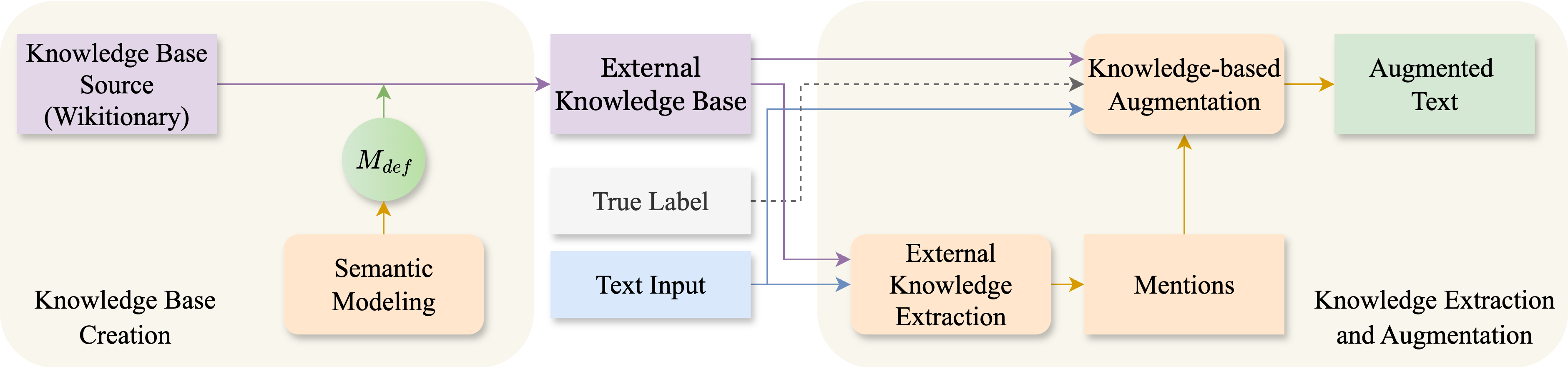}
    \caption{Overview of external knowledge base preparation and input data augmentation. The data flow relating to the dotted line applies only to semantic augmentation, as discussed in Section~\ref{sec:augmentation}.}
    \label{fig:augmentation-approach}
\end{figure*}

\subsection{Knowledge Base Creation}
\label{sec:knowledge-base-creation}

Figure~\ref{fig:augmentation-approach} illustrates different components of the augmentation process given an instance and the knowledge base. 

\subsubsection{Knowledge Base Source}

The knowledge base used in this work needs to have several characteristics and specifications. We assume that the knowledge base is represented as a graph (i.e., knowledge graph) with relationships between entities represented as triples (i.e., $<subject,~predicate,~object>$ tuple). The \textit{subject} and \textit{object} are nodes in the graph, and each node represents a specific sense of a word. Senses are defined by a gloss - an English definition of the sense of the word~\cite{miller1990introduction}. 

The above requirements are satisfied in the Wiktionary~\footnote{https://www.wiktionary.org/}~\cite{wiktionary} and WordNet~\footnote{https://wordnet.princeton.edu/}~\cite{miller1990introduction} knowledge bases.  Compared to WordNet, which mainly contains formal vocabulary and definitions, Wiktionary includes crowd-sourced knowledge contributed by its online community of users that often encompasses slang and colloquial expressions. This community-driven model allows Wiktionary to reflect emerging language trends and update content more rapidly than expert-curated resources like WordNet. As a result, Wiktionary is particularly valuable for applications of online behavioral analytics with tasks such as hate speech classification, where timely recognition of informal and evolving language is essential. Therefore, we selected Wiktionary as the knowledge base. However, Wiktionary is a semi-structured data source and cannot be directly utilized in machine learning models. Hence, we used \textit{Wiktextract}, a resource that has already converted its semi-structured data to a structured format~\cite{ylonen2022wiktextract}. We further processed the data from \textit{Wiktextract} to suit the needs of the proposed approach and represented it with knowledge triples. We extracted all the words from Wiktionary and then identified their senses. Then, for each sense  (\textit{subject}), we extracted the relationships (\textit{predicate}) to other words (\textit{object}). We extracted only the relations of types: forms, synonyms, hyponyms, and instances for simplicity. 

\subsubsection{Semantic Modeling}

Random augmentation through the replacement of words with synonyms or related terms could prove ineffective due to semantic mismatches. Semantic modeling~\cite{guo2022semantic} aims to enhance augmentation by generating contextually meaningful variations of the original text. For example, the term ``Oreo'' could be substituted with ``race traitor'' based on definitions in contexts involving hate speech. In contrast, a substitution such as ``cookie'' may be more appropriate for straightforward augmentation in more neutral contexts. This highlights the importance of contextual understanding in selecting semantically aligned terms during data augmentation. 
A commonly used augmentation technique in the literature involving random replacement, in which observed terms are substituted with related terms from a knowledge base, can potentially result in inaccurate augmentations. 
In general knowledge bases, meaning is typically encoded through distinct senses associated with the same lexical form, each accompanied by a gloss or definition. Resources such as Wiktionary explicitly represent this definition-based semantic relation.  

Note that the availability of a relevant definition of a sense alone is not enough; we need to be able to identify if the word in context (sense of the word) is related to the target label for semantic augmentation based on the available information (i.e., definitions). For this purpose, we introduce the semantic modeling task, which is distinct from the final task for deviant behavior classification. In the semantic modeling task, a model is provided with a definition ($def(W_{s})$) of a sense of a word ($W_s$) that outputs whether the provided definition is relevant to a behavior of interest (e.g., hate speech or not). To train such a model, we fine-tune ``sentence-transformers/all-MiniLM-L6''~\cite{reimers-2019-sentence-bert} model using the SetFit library~\footnote{https://huggingface.co/docs/setfit/en/index} (producing a SetFit model). 

For training this model, we created a variant of the HateWIC dataset~\cite{hoeken2024hateful} with some modifications to the data structure. HateWIC dataset contains the following fields: text, word, the in-context definition of the word, and label (hateful/normal). We selected only the unique definitions from that dataset and counted the number of times they appeared in hateful and non-hateful contexts. We obtained the final label per definition by taking the majority label. Then, we validated the performance of this model on a separate validation set and observed that it achieved an accuracy of $80\%$. Next, we used this model to predict the relevance of definitions in Wiktionary to deviant behavior related to hate speech.


\subsection{Knowledge Extraction and Augmentation}
\label{sec:augmentation}

\textbf{External Knowledge Extraction}: External knowledge extraction constitutes a core component of our approach, enabling augmentation. This module processes text and identifies entries in the knowledge base relevant to the text (known as Mentions) via a lexical-matching algorithm. To accelerate matching, we employ a trie data structure that efficiently retrieves all forms of each word in the text. These variants are sourced directly from the ``forms'' relations in the Wiktionary knowledge base. At this stage, it remains to determine whether the extracted entries correspond to the intended sense within the given context.  

\textbf{Knowledge-based Augmentation}: 
We perform augmentation by replacing text spans with the following predicates found in the knowledge base: ``synonym", ``hyponym'', and ``instance''. We use two main strategies for testing the augmentation process: 1) random replacement and 2) semantic replacement with the help of the definition model.

\noindent\textbf{\textit{Random Augmentation}}: The random augmentation involves replacing text spans that match with a knowledge base entity by using a random relation of that entity.

\noindent
\textbf{\textit{Semantic Augmentation}}: This involves augmenting specific instances labeled with deviant behavior classes, i.e., ``hateful'' or ``offensive'', using entity relations in the knowledge base that have a subject with a hate speech-related definition. For instances labeled with other classes, we perform augmentation using relations that have subjects with non-hate speech-related definitions. 

The above approaches were performed on data in the memory buffer (from previous tasks) before the learning process and the current training dataset before the selection process. 
Random augmentation was performed as an alternative to compare the performance. Moreover, we only augment an instance with a given triple with probability \textbf{p} to ensure that we generate a limited number of augmented instances. We selected \textbf{p} such that we have approximately a similar number of augmentation instances compared to semantic augmentation. 

\subsection{Memory Buffer and Exemplar Selection}
\label{sec:method-exempar-selection}

The memory buffer stores representative instances, or exemplars, from the new task training data to be used later to prevent catastrophic forgetting. Typically, the buffer is updated by selecting representative instances from the training data of the new task and the data already stored in memory while preserving total memory size constraints. The selection is performed after the system has been trained on the new task. In prior work, one promising approach is cluster-based selection~\cite{qian2021lifelong}. Cluster-based selection involves creating $\textit{k} (= M / t)$ clusters from task data after training on task $\tau_t$ and selecting the centroids to be saved in the buffer. In our proposed method, we employed cluster-based selection using the \textit{k}-means algorithm. We further improved this cluster-based selection approach with the help of a knowledge base by first augmenting the training data and then applying clustering. 

\begin{table*}[!th]
\centering
\caption{Summary of datasets used for this study. The ``Original Size'' column indicates the number of instances prior to restructuring for CL tasks.}
\label{tab:dataset-overview}
\begin{tabular}{llllc}
\toprule
\textbf{ID} & \textbf{Dataset} & \textbf{Domain / Platform}  & \textbf{Original Size} & \textbf{\# Tasks} \\
\midrule
$\mathcal{D}_B$ & Civil Comments (2019)~\cite{borkan2019nuanced} & News comments & 1,999,516 & 5 \\
$\mathcal{D}_H$ & HateXplain (2021)~\cite{mathew2021hatexplain} & Twitter, Gab & 20,148 & 5 \\
$\mathcal{D}_K$ & Kennedy et al. (2020)~\cite{kennedy2020constructing} & YouTube, Twitter, Reddit & 39,565 & 5 \\
\bottomrule
\end{tabular}
\end{table*}

\begin{table}[!ht]
    \centering
    \caption{Summary of tasks from datasets in the Table~\ref{tab:dataset-overview}. The tasks are presented in the same order as they were presented to the CL system. The task order is randomly selected. \label{tab:task-overview}}
    \begin{tabular}{llcccc}
        \toprule
        DS  & Task  & Train & Valid & Test & \#Classes \\
        \midrule
        \multirow[c]{5}{*}{$\mathcal{D}_B$} & \textbf{Religion} & 4000  & 200 & 400  & 2 \\
        \textbf{} & \textbf{Gender} & 4000  & 200 & 400  & 2 \\
        \textbf{} & \textbf{Race ethnicity} & 4000 & 200 & 400  & 2 \\
        \textbf{} & \textbf{Sexuality} & 3238 & 200 & 400  & 2 \\
        \textbf{} & \textbf{Disability} & 2557  & 200 & 400  & 2 \\
        \cline{1-6}
        \multirow[c]{5}{*}{$\mathcal{D}_H$} & \textbf{Refugee} & 99 & 100 & 300  & 2 \\
        \textbf{} & \textbf{Homosexual} & 410 & 100 & 300  & 2 \\
        \textbf{} & \textbf{Muslim} & 272 & 150 & 450  & 3 \\
        \textbf{} & \textbf{Women} & 560 & 100 & 300  & 2 \\
        \textbf{} & \textbf{African} & 1062 & 150 & 450  & 3 \\
        \cline{1-6}
        \multirow[c]{5}{*}{$\mathcal{D}_K$} & \textbf{Religion} & 1692  & 300 & 600  & 3 \\
        \textbf{} & \textbf{Origin} & 1367 & 300 & 600  & 3 \\
        \textbf{} & \textbf{Race} & 3269 & 300 & 600  & 3 \\
        \textbf{} & \textbf{Sexuality} & 1447 & 300 & 600  & 3 \\
        \textbf{} & \textbf{Gender} & 6000  & 300 & 600  & 3 \\
        \cline{1-6}
        \bottomrule
    \end{tabular}
\end{table}

\begin{table*}[ht]
\centering
\caption{Experiment results for different approaches of naive and direct replay settings. \textbf{NF}, \textbf{RD}, \textbf{SR}, \textbf{CS} indicate Naive Fine-tuning, Random Direct Replay, Stratified Random Direct Replay, and cluster-based selection for replay. \textbf{KR} represents the approach with proposed augmentation with the external knowledge base before the selection process and learning. The random and semantic augmentation variants of this approach are included in the results. The values in the table are represented as percentages. The \textbf{bold} and \underline{underlined} values indicate the best and the second-best performance values, respectively.}
\label{tab:experiment_results}
\begin{tabular}{l|cc|cc|cc}
\toprule
& \multicolumn{2}{c}{$\mathcal{D}_B$} 
& \multicolumn{2}{c}{$\mathcal{D}_H$} & \multicolumn{2}{c}{$\mathcal{D}_K$} \\
Approach & A & AUC & A & AUC & A & AUC \\
\midrule
\textbf{NF} & $12.9^{\pm2.0}$ & $61.0^{\pm2.2}$ & $6.7^{\pm0.0}$ & $52.7^{\pm1.5}$ & $11.8^{\pm0.3}$ & $59.4^{\pm5.6}$ \\
\textbf{RD} & $64.2^{\pm6.0}$ & $83.7^{\pm0.4}$ & $15.5^{\pm4.7}$ & $54.9^{\pm0.5}$ & $25.7^{\pm6.4}$ & $70.1^{\pm0.2}$ \\
\textbf{SR} & $65.1^{\pm5.0}$ & $83.5^{\pm0.1}$ & $16.7^{\pm5.2}$ & $57.2^{\pm3.6}$ & $27.8^{\pm4.6}$ & $70.5^{\pm0.7}$ \\
\textbf{CS} & $64.6^{\pm3.2}$ & $83.0^{\pm0.4}$ & $15.4^{\pm5.7}$ & $56.9^{\pm3.6}$ & $24.8^{\pm6.4}$ & $68.8^{\pm5.5}$ \\ \hline
\textbf{KR$_{rnd}$ (proposed)} & $\underline{75.9^{\pm0.1}}$ & $\underline{85.6^{\pm0.6}}$ & $\underline{46.2^{\pm1.6}}$ & $\boldsymbol{63.5^{\pm1.2}}$ & $\underline{43.3^{\pm0.8}}$ & $\boldsymbol{74.3^{\pm0.4}}$ \\ 
\textbf{KR$_{sem}$ (proposed)} & $\boldsymbol{76.2^{\pm1.2}}$ & $\boldsymbol{85.8^{\pm0.2}}$ & $\boldsymbol{46.7^{\pm1.9}}$ & $\underline{63.4^{\pm1.1}}$ & $\boldsymbol{43.4^{\pm1.7}}$ & $\underline{73.7^{\pm0.5}}$ \\
\bottomrule
\end{tabular}
\end{table*}

\section{Experiments}
\label{sec:experiments}

\subsection{Data}

Our primary criterion for selecting corpora for continual deviant behavior classification was the presence of an attribute that would allow us to partition the data into discrete tasks. We surveyed several sources for deviant behavior analysis in online data, including~\cite{vidgen2020directions,vidgen2021introducing,ousidhoum2019multilingual,yoder2022hate}, and the Hate Speech Dataset Catalogue~\footnote{https://hatespeechdata.com}. We identified multiple candidate datasets from these sources. We used the ``target'' attribute for deviant behavior to define each task, so only the datasets that included this attribute were retained. The target indicates the category of identities, such as religion, in the context of hate speech. In addition, we discarded instances annotated with multiple targets to simplify task separation. Non-English datasets were also excluded from this research due to the low availability of datasets. Finally, datasets with fewer than 200 instances per class (with at least five tasks) were deemed too small for effective fine-tuning of a language model and CL-based evaluation and were therefore excluded.

Table~\ref{tab:dataset-overview} presents a summary of datasets in each CL task created for our experiments. Note that we randomly select 2000 instances from the training data if the dataset size exceeds 2000 data points to mitigate challenges related to resource requirements when executing the experiments. Table~\ref{tab:task-overview} illustrates the distribution of classes for each task.

\subsection{Experimental Settings}
\label{sec:experiment-settings}

We experiment with the following CL approaches to observe the impact of the proposed knowledge integration method on CL system performance. 

\textbf{Na\"ive Fine-tuning (NF)}: Na\"ive is just a model that is being fine-tuned on the new data as it is observed by the system. There will be no knowledge (internal or external) other than knowledge transfer from using the same parameters across the tasks (transfer learning). This is a \underline{baseline} approach used for comparison.

\textbf{Direct Replay}: Direct replay is an additional set of \underline{baseline} approaches for comparison. These methods use a fixed-size memory to store instances from each task, which can be helpful in future learning tasks. We experimented with three variants of this approach with different selection strategies as follows:

\begin{enumerate}
    \item \textit{Random Selection (\textbf{RD})}: We randomly choose the same number of instances from each new task. 
    \item \textit{Stratified Random (\textbf{SR})}: We randomly choose an equal number of instances from each class label within every task.
    \item \textit{Cluster-based Selection (\textbf{CS})}: For each task, we first cluster its documents, then select the same number of instances from each resulting cluster. Prior work has shown that this method outperforms simple random sampling by identifying and retaining instances that are more distinct and informative in the distributional space.
\end{enumerate}

\textbf{Knowledge-aware Direct Replay (\textbf{KR}):} Our \underline{proposed} approach that integrates external knowledge into the direct replay approach.
There are two variants of this as follows:
\begin{enumerate*}
    \item Random Augmentation (\textbf{KR}$_{rnd}$);
    \item Semantic Augmentation (\textbf{KR}$_{sem}$);
\end{enumerate*}.

\subsection{Evaluation Measures}

\begin{table*}[ht]
\centering
\caption{The forgetting metric for each approach, dataset, and metric. Lower is better. The symbols indicating approaches have the same meaning as in Table~\ref{tab:experiment_results}.}
\label{tab:fgt_results}
\begin{tabular}{l|cc|cc|cc}
\toprule
& \multicolumn{2}{c}{$\mathcal{D}_B$} 
& \multicolumn{2}{c}{$\mathcal{D}_H$} & \multicolumn{2}{c}{$\mathcal{D}_K$} \\
Approach & $\mathcal{F}_{A}$ & $\mathcal{F}_{AUC}$ & $\mathcal{F}_{A}$ & $\mathcal{F}_{AUC}$ & $\mathcal{F}_{A}$ & $\mathcal{F}_{AUC}$ \\
\midrule
\textbf{NF} & $77.3^{\pm3.4}$ & $32.9^{\pm2.9}$ & $46.1^{\pm0.4}$ & $3.0^{\pm2.0}$ & $49.4^{\pm6.6}$ & $21.3^{\pm6.3}$ \\
\textbf{RD} & $17.3^{\pm6.7}$ & $5.0^{\pm1.1}$ & $30.7^{\pm6.9}$ & $3.0^{\pm1.3}$ & $35.4^{\pm5.5}$ & $8.3^{\pm1.6}$ \\
\textbf{SR} & $16.0^{\pm6.2}$ & $5.2^{\pm0.7}$ & $31.8^{\pm6.1}$ & $2.9^{\pm1.4}$ & $34.2^{\pm4.7}$ & $8.7^{\pm1.1}$ \\
\textbf{CS} & $15.9^{\pm3.9}$ & $5.8^{\pm0.7}$ & $32.6^{\pm6.8}$ & $2.8^{\pm1.6}$ & $34.7^{\pm5.1}$ & $8.8^{\pm4.6}$ \\ \hline
\textbf{KR$_{rnd}$ (proposed)} & $\underline{4.2^{\pm0.9}}$ & $\underline{2.7^{\pm0.6}}$ & $\underline{1.0^{\pm0.4}}$ & $\underline{0.3^{\pm0.2}}$ & $\boldsymbol{13.4^{\pm3.9}}$ & $\boldsymbol{3.4^{\pm0.8}}$ \\
\textbf{KR$_{sem}$ (proposed)} & $\boldsymbol{3.5^{\pm0.6}}$ & $\boldsymbol{2.3^{\pm0.4}}$ & $\boldsymbol{0.7^{\pm0.6}}$ & $\boldsymbol{0.0^{\pm0.1}}$ & $\underline{14.1^{\pm4.0}}$ & $\underline{4.6^{\pm1.0}}$ \\
    \bottomrule
\end{tabular}
\end{table*}

We employ two primary evaluation methods commonly used in the CL literature: average performance and forgetting. We calculate performance in terms of standard metrics, including accuracy and the Area Under the ROC Curve (AUC) score, as these provide a fair comparison between CL tasks. The AUC score is calculated class-wise and averaged separately for each task. 

Additionally, we evaluate the forgetting metric ($\mathcal{F_P}$), where $P$ represents the performance metric (accuracy and AUC score) used to calculate the forgetting of a model. This metric is calculated as the average reduction in model performance~\cite{zhou2024class}, measured by the difference between the highest performance achieved for each task and the performance observed after training in the final task. 

\subsection{Implementation Details}

We used the BERT-Base model as the initial pre-trained transformer-based model from the HuggingFace library~\footnote{https://huggingface.co/google-bert/bert-base-uncased}. We determined the training hyperparameters for a single hate speech classification task (the first task of $\mathcal{D}_K$) using the Optuna library~\footnote{https://optuna.org/}. 
We evaluated models with different learning rates, batch sizes, weight decay, and early stopping patience values, and finally settled on 20 epochs with early stopping with a patience of 8 epochs. 
The learning rate was set to $5e^{-6}$ with a batch size of $16$. Moreover, we use 0.1 weight decay during training. We evaluated with memory size ($M$) of $500$ instances where applicable, and we use random augmentation probability (\textbf{p}) value of 15\%. We have set the value of $\lambda$ to 1 to treat all instances equally during training. 

\section{Result Analysis and Discussion}
\label{sec:results}

\begin{figure*}[!h]
    \centering
    \includegraphics[width=0.6\linewidth]{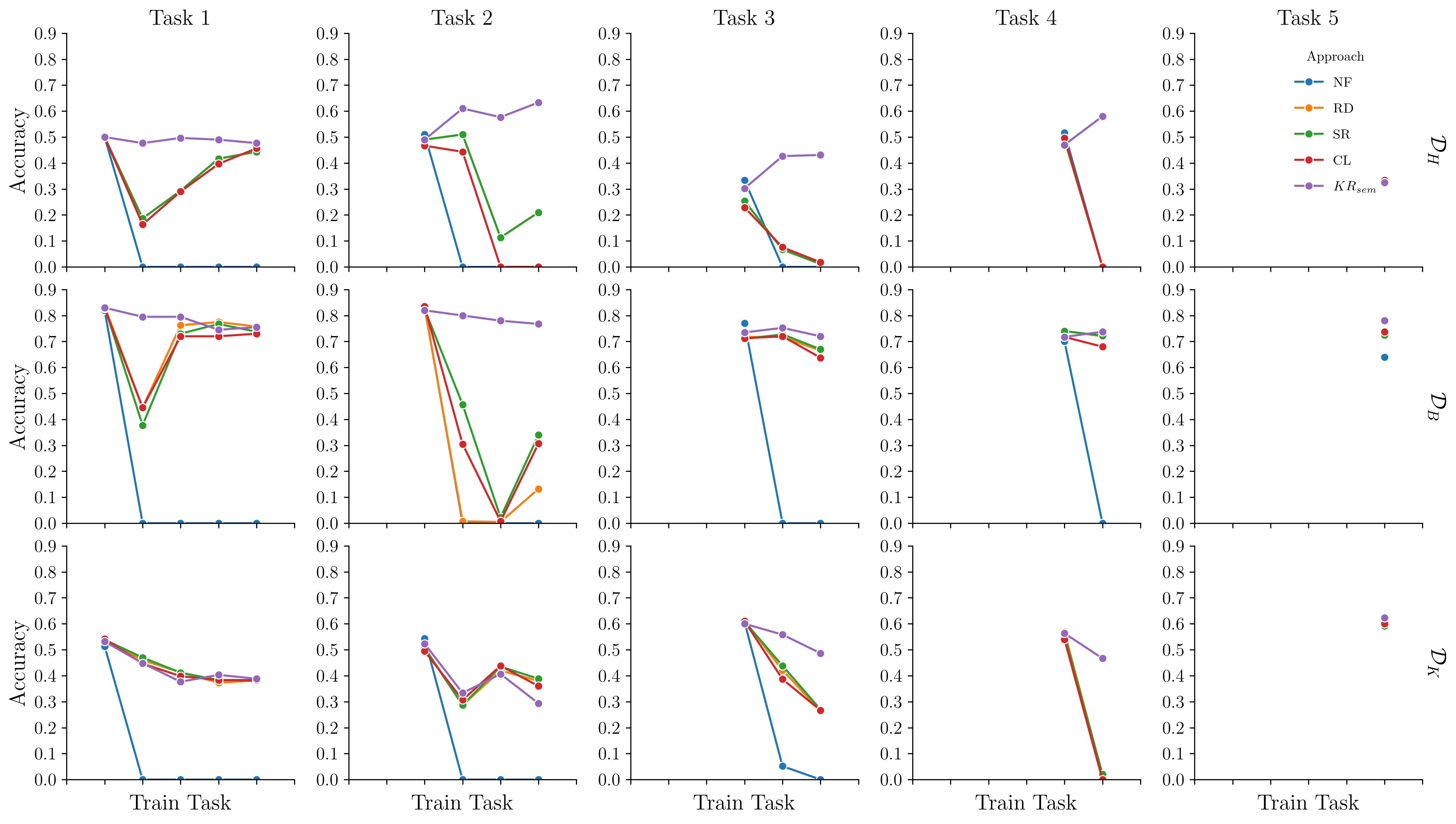}
    \caption{Accuracy of individual tasks at each task training stage. The x-axis represents the training task stage, and the y-axis represents the accuracy. Each plot in the column and row corresponds to a test task and dataset, respectively. The tasks are numbered in the same order as they appear for training.}
    \label{fig:accuracy-cl}
\end{figure*}

\subsection{Prediction Performance for Knowledge-guided Approach} 

We first analyzed the performance in the last task and the forgetting of the model between the proposed augmentation approach and the baseline approaches. Specifically, we considered four other approaches for the comparison as follows: 
1) na\"ive fine-tuning approach (\textbf{NF}), 
2) direct replay with random selection (\textbf{RD}), 
3) direct replay with stratified selection (\textbf{SR}), 
4) cluster-based direct replay (\textbf{CS}), and 
5) augmentation-based direct replay (our approach \textbf{KR$_x$}). 

\subsubsection{Average Performance Scores}

Table~\ref{tab:experiment_results} illustrates the performance in terms of average accuracy (A) and AUC scores after the last task across all the observed tasks in each of the four datasets. We find that our proposed augmentation approach outperforms all existing approaches in the replay-based CL setting when using the same buffer memory size. This finding supports our proposal of using external knowledge.

Overall, the minimum performance is observed in the \textbf{NF} approach, which is expected since it does not employ any specific strategy to address the core challenge of replay-based CL settings. Furthermore, among the three selection strategies for direct relay approaches without external knowledge, we observe that the \textbf{SR} method outperforms the other two in most cases. Interestingly, the lower performance of the \textbf{CS} method contradicts prior studies demonstrating the effectiveness of cluster-based selection strategies~\cite{qian2021lifelong}. We may attribute the difference in results partly to the combined differences in the nature of tasks across datasets and the specific algorithms used for clustering. Additionally, we observe that the proposed \textbf{KR} methods outperformed other methods, with \textbf{KR}$_{sem}$ marginally outperforming \textbf{KR}$_{rand}$ in terms of the accuracy metric. The reason why the AUC score has higher performance for the \textbf{KR}$_{rand}$ could be due to the threshold-independent nature of the metric. We can further ascertain that our proposed \textbf{KR}$_{sem}$ method outperforms the baseline approaches by comparing the performance of each task as indicated in Figure~\ref{fig:accuracy-cl}, where it shows a higher performance of tasks at the end of the CL process for our proposed method.

\subsubsection{Forgetting Scores}

Table~\ref{tab:fgt_results} presents the forgetting scores for the same set of approaches described in Table~\ref{tab:experiment_results}. We observe similar patterns in the forgetting scores as we did in the average performance results. The most performance drop has been observed in the \textbf{NF} approach since we did not employ any specific strategies to prevent catastrophic forgetting. There is no clear winner among the three direct replay approaches that do not use external knowledge when compared across all datasets. This implies that the characteristics of the datasets and the tasks may affect the performance. We observe that both \textbf{KR} methods preserve most knowledge compared to other methods, with \textbf{KR}$_{sem}$ marginally outperforming \textbf{KR}$_{rand}$ except in $\mathcal{D}_K$. 
We explore the possible causes for this marginal improvement in Section~\ref{sec:augmentation_analysis} with an example.
This relationship shows that the performance loss across all tasks is, on average, minimal in the proposed augmentation-based approaches. 

Based on both the performance score values and the forgetting scores, we can conclude that \textbf{integrating external knowledge through augmentation in replay-based continual learning reduces catastrophic forgetting and enhances overall system performance, thereby addressing RQ1.} 

\subsection{Ablation Study}

We examine the effectiveness of using augmented training data at different stages of the proposed knowledge-guided continual learning framework and compare them across datasets using the last task performance. 
\begin{table*}[!ht]
\centering
\caption{
Results of the ablation study. The first row illustrates the performance of our \textbf{KR}$_{sem}$ approach, and the following two lines indicate the performance of the approach without augmentation at two distinct stages where augmentation is used. The best results are in \textbf{bold}, and the next best results are \underline{underlined}.
\label{tab:ablation_results}} 
\begin{tabular}{lcccccc}
\toprule
& \multicolumn{2}{c}{$\mathcal{D}_B$} & \multicolumn{2}{c}{$\mathcal{D}_H$} & \multicolumn{2}{c}{$\mathcal{D}_K$} \\
&  A & AUC & A & AUC & A & AUC \\
\midrule
\textbf{Full Model} & $\boldsymbol{76.2^{\pm1.2}}$ & $\boldsymbol{85.8^{\pm0.2}}$ & $\boldsymbol{46.7^{\pm1.9}}$ & $\boldsymbol{63.4^{\pm1.1}}$ & $\boldsymbol{43.4^{\pm1.7}}$ & $\boldsymbol{73.7^{\pm0.5}}$ \\
\textbf{w/o Pre-selection Augmentation} & $\underline{75.5^{\pm0.5}}$ & $\underline{85.3^{\pm0.3}}$ & $\underline{45.6^{\pm0.9}}$ & $\underline{62.4^{\pm0.6}}$ & $\boldsymbol{43.4^{\pm2.1}}$ & $\underline{73.5^{\pm0.4}}$ \\
\textbf{w/o Pre-learning Augmentation} & $63.3^{\pm1.0}$ & $82.8^{\pm0.7}$ & $16.9^{\pm1.3}$ & $55.9^{\pm2.0}$ & $\underline{27.2^{\pm2.8}}$ & $69.9^{\pm0.6}$ \\
\bottomrule
\end{tabular}
\end{table*}
Table~\ref{tab:ablation_results} showcases the performance of our approach without augmentation of the new training dataset before selection (pre-selection augmentation) and without augmentation of data in the buffer before the learning process (pre-learning augmentation) separately. We observe consistent performance drops across all datasets when augmentation is removed from the continual learning system during both pre-selection and pre-learning stages. Additionally, we observe a larger drop in performance when we learn without augmentation in the pre-learning stage across all datasets. This implies that the augmentation does contribute relevant task knowledge to improve the learning process. \textbf{We can conclude that the augmentation-based knowledge integration in replay-based CL approaches is effective when performed during both the selection and learning phases of CL, and thereby, we address RQ2}.

\subsection{Augmentation Analysis}
\label{sec:augmentation_analysis}

The marginal performance improvement of the semantic approach could be attributed to the semantic augmentation of the instances instead of random augmentation. The reason it may only be a marginal improvement is that there is a smaller number of words that are ambiguous given the context, or when an instance with deviant behavior is present, only a few words in it are actually related to hate speech, while others are generic. Generic words can be easily replaced/augmented without any additional knowledge~\cite{pellicer2023data}. We conducted a qualitative analysis of several augmentation examples to illustrate this point further.

\begin{figure}[!h]
    \centering
    \includegraphics[width=0.9\linewidth]{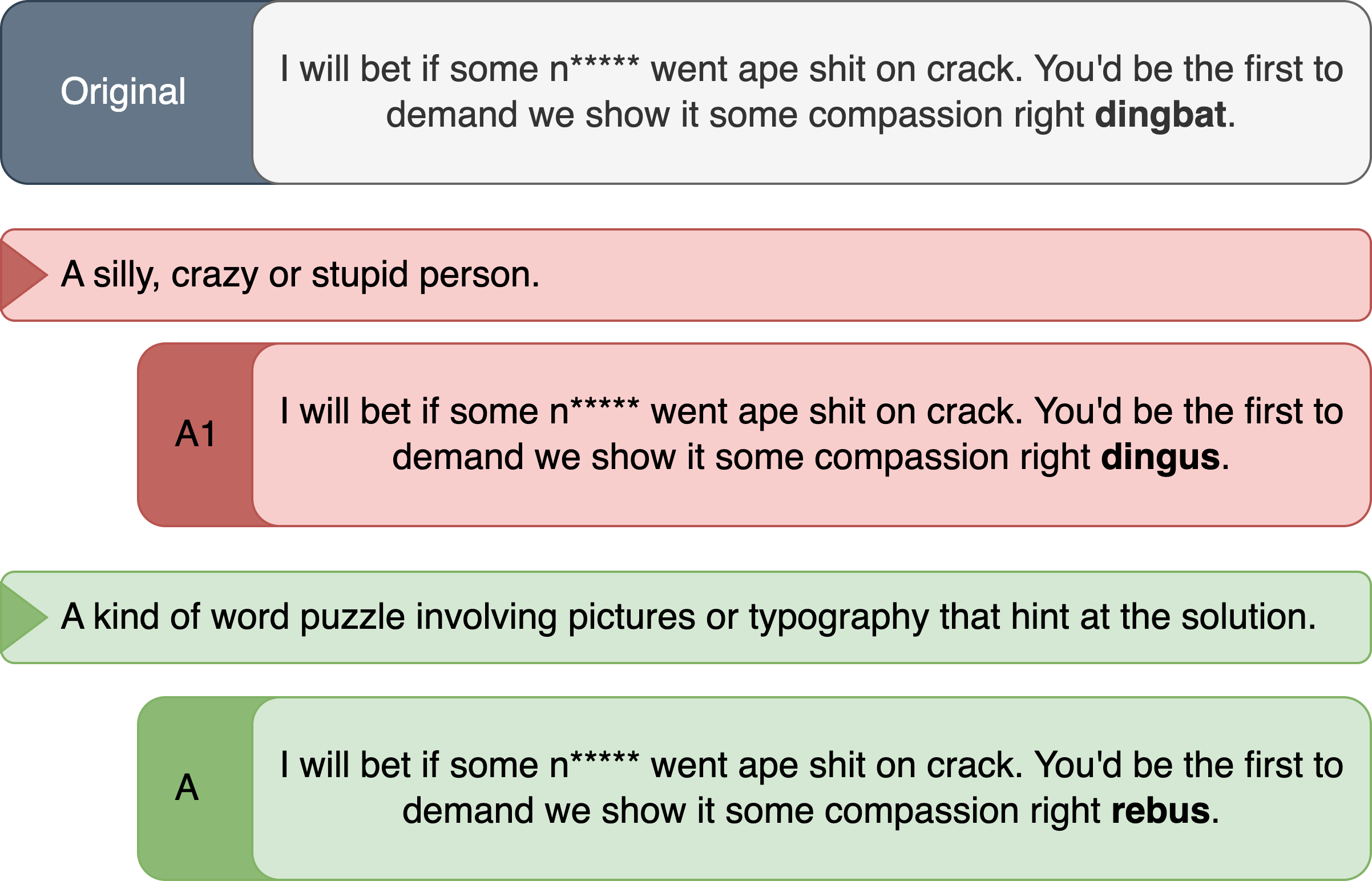}
    \caption{An augmentation example. Illustrates both semantic augmentation (A1) and incorrect augmentation (A2).}
    \label{fig:example-augmentations}
    \vspace{-0.3in}
\end{figure}

Figure~\ref{fig:example-augmentations} illustrates an example from our experimental data. The highlighted word `\textbf{dingbat}' has multiple senses, according to Wiktionary~\footnote{https://en.wiktionary.org/wiki/dingbat}. Two of these senses are noted in the figure, with one meaning relating to divergent behaviors and the other not. In our semantic augmentation approach, only one that relates to divergent behavior is generated (A1) since the original message is labeled as divergent in the dataset. However, this means that any other word in the same message that is not divergent will not be augmented at all when we are using the semantic augmentation strategy. Therefore, the number of candidate augmentations generated using the semantic augmentation approach is limited, whereas when we use random augmentation, this is not the case. Random augmentation, however, will create augmentations like the second augmentation example (A2) in the figure, which is not semantically correct. 

\begin{figure*}[!th]
    \centering
    \includegraphics[width=0.6\linewidth]{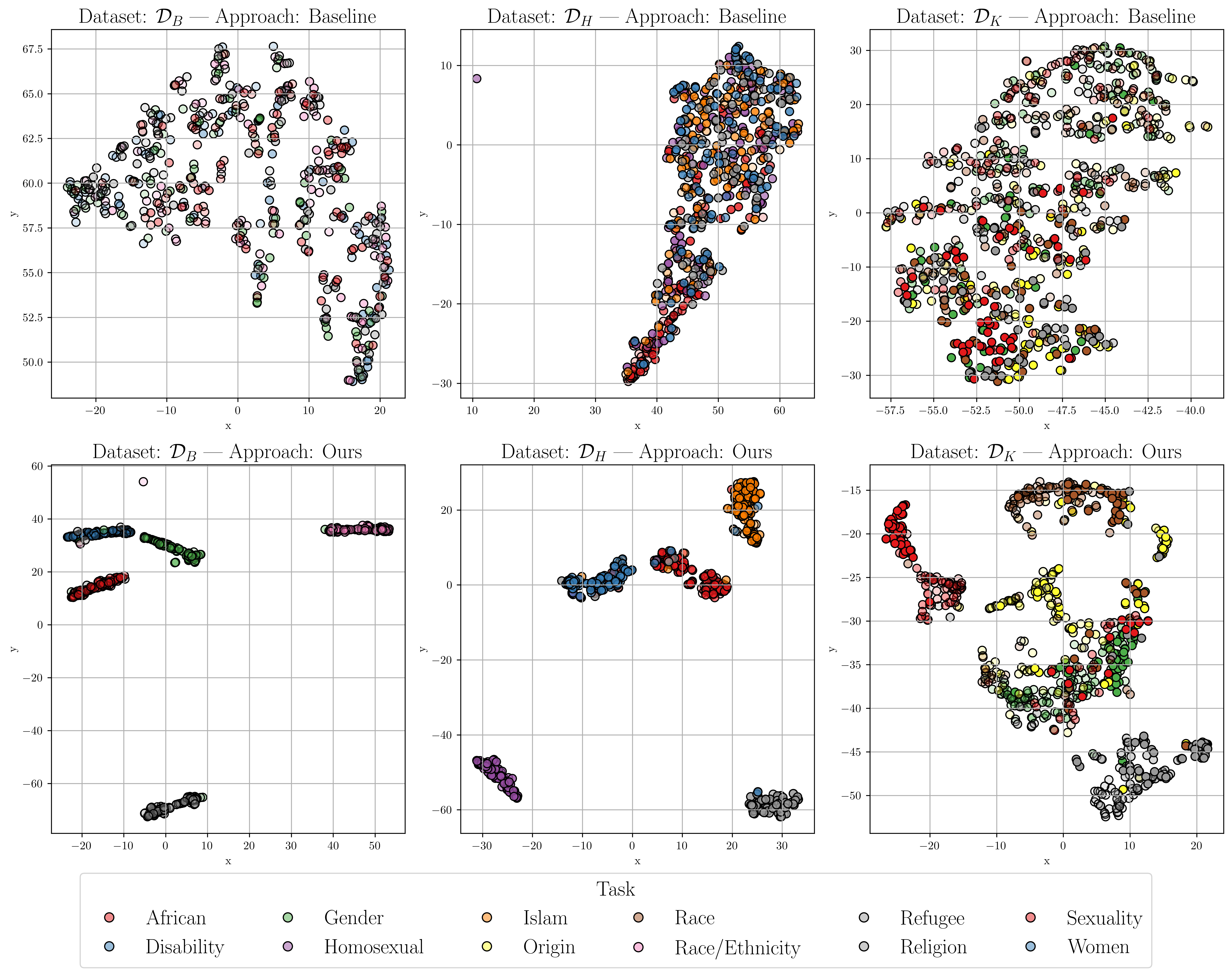}
    \caption{The t-SNE [13] visualization of the last layer of the BERT model after training on all tasks of each dataset using the baseline model (\textbf{NB}) and our model (\textbf{KR}$_{sem}$).}
    \label{fig:embeddings}
\end{figure*}

Moreover, we provide a t-SNE~\cite{van2008visualizing} visualization of the computed feature representations of the fine-tuned BERT model on three datasets in Figure~\ref{fig:embeddings} for the baseline approach (\textbf{NF}) and our approach (\textbf{KR}$_{sem}$). For each instance, we compute the embedding representation by taking the vector corresponding to the [CLS] token in the last layer of the BERT model. The colors in the plot indicate tasks. We notice that the \textbf{KR}$_{sem}$ has been able to learn about the tasks considerably better than the baseline model across all datasets. This indicates that the learned feature space was discriminative of the tasks observed throughout the CL process, which is likely the reason why the proposed approaches achieved the best performance.

\section{Conclusion and Future Work}
\label{sec:conclusion}

In this paper, we proposed an external knowledge base-guided approach to enhance the CL process for behavioral analytics on online data that often contains evolving user behavior. To achieve this, we introduce a novel external knowledge component and investigate diverse strategies for incorporating external knowledge into the traditional CL framework. Specifically, we use data augmentation in replay-based CL approaches by enhancing the selection process for data instances before updating the internal memory and then augmenting the exemplars in the memory buffer during the data replay phase. Moreover, we demonstrate that the effectiveness of augmentation can be further improved by employing strategies such as semantic augmentation, which improves continual learning performance while maintaining a constant buffer size. Using three datasets from prior research on deviant behavior analysis, particularly hate speech, we conducted extensive experimental analysis to demonstrate that integrating external knowledge through augmentation in replay-based continual learning helps reduce catastrophic forgetting and enhance overall system performance.

\textbf{Limitations \& Future Work}: 
In our study, we incorporate external knowledge through augmentation, which occurs outside the model fine-tuning stage. Additional techniques for infusing knowledge, such as integrating it as input features or applying learning constraints, could further enhance model learning~\cite{sheth2022defining}.
In our experiments, tasks were divided solely by identity group. While this design simplifies evaluation, it may be less practical in real-world scenarios, where distribution shifts can occur even within data from the same identity. Future research could investigate more granular task divisions to address this limitation.
Our approach and experiments focus on classifying English-language text related to deviant behaviors. Further studies should explore whether the proposed method can yield similar CL improvements in other languages and task domains.
Wiktionary, used as our primary external knowledge source, is a general-purpose resource curated by volunteers. Although it benefits from policies and algorithms that limit adversarial edits~\footnote{https://en.wiktionary.org/wiki/Wiktionary:Policies\_and\_guidelines}, its quality may be lower than that of expert-curated, domain-specific knowledge bases. Future work could incorporate additional resources, such as ConceptNet~\footnote{https://conceptnet.io/}, and explore combining multiple knowledge sources to strengthen the augmentation process.
Finally, the proposed approach may incur additional computational overhead during the augmentation process. When deploying the system, the cost of this extra processing should be carefully weighed against the available resources.

\section*{Acknowledgment}

Authors acknowledge the partial support of research grants from the Western Norway Research Institute's SOCYTI project, grant \# 331736, and INTPART DTRF project, grant \# 309448. 
This project was supported by resources provided by the Office of Research Computing at George Mason University (URL: https://orc.gmu.edu), which was funded in part by grants from the National Science Foundation (Award Number 2018631).

\bibliographystyle{IEEEtran}
\bibliography{references}

%

\end{document}